\documentclass[journal,twoside,web]{ieeecolor}
\usepackage{generic}
\usepackage{cite}
\usepackage{amsmath,amssymb,amsfonts}
\usepackage{algorithmic}
\usepackage{graphicx}
\usepackage{algorithm,algorithmic}
\usepackage{hyperref}
\hypersetup{hidelinks=true}
\usepackage{textcomp}
\usepackage{booktabs}
\usepackage{multirow}
\usepackage{xcolor}
\usepackage[table]{xcolor}

\def\BibTeX{{\rm B\kern-.05em{\sc i\kern-.025em b}\kern-.08em
    T\kern-.1667em\lower.7ex\hbox{E}\kern-.125emX}}
\begin{document}
\title{ChemFixer: Correcting Invalid Molecules to Unlock Previously Unseen Chemical Space}
\author{Jun-Hyoung Park, Ho-Jun Song, and Seong-Whan Lee, \IEEEmembership{Fellow, IEEE}
\thanks{This work was supported by the Institute of Information \& communications Technology Planning \& Evaluation (IITP) grant funded by the Korea government (MSIT) (No. RS-2019-II190079, Artificial Intelligence Graduate School Program (Korea University)).}
\thanks{Jun-Hyoung Park, Ho-Jun Song, and Seong-Whan Lee are with the 
Department of Artificial Intelligence, Korea University, Seongbuk-ku, Seoul   02841, Republic of Korea (e-mail: junhyoung\_park@korea.ac.kr; h\_j\_song@korea.ac.kr; sw.lee@korea.ac.kr).}}

\maketitle

\begin{abstract}
Deep learning–based molecular generation models have shown great potential in efficiently exploring vast chemical spaces by generating potential drug candidates with desired properties. However, these models often produce chemically invalid molecules, which limits the usable scope of the learned chemical space and poses significant challenges for practical applications. To address this issue, we propose ChemFixer, a framework designed to correct invalid molecules into valid ones. ChemFixer is built on a transformer architecture, pre-trained using masking techniques, and fine-tuned on a large-scale dataset of valid/invalid molecular pairs that we constructed. Through comprehensive evaluations across diverse generative models, ChemFixer improved molecular validity while effectively preserving the chemical and biological distributional properties of the original outputs. This indicates that ChemFixer can recover molecules that could not be previously generated, thereby expanding the diversity of potential drug candidates. Furthermore, ChemFixer was effectively applied to a drug–target interaction (DTI) prediction task using limited data, improving the validity of generated ligands and discovering promising ligand–protein pairs. These results suggest that ChemFixer is not only effective in data-limited scenarios, but also extensible to a wide range of downstream tasks. Taken together, ChemFixer shows promise as a practical tool for various stages of deep learning–based drug discovery, enhancing molecular validity and expanding accessible chemical space.
\end{abstract}


\section{Introduction}
\label{sec:introduction}
\IEEEPARstart
{M}{achine} learning has been widely applied in biomedical and healthcare domains, driving research progress. These applications encompass various areas, including disease, infection, and biomarker research, as well as digital healthcare and preventive medicine \cite{add4}, \cite{add3}, \cite{add5}. Within this broad context, drug design and molecular generation are regarded as prominent application areas \cite{add1}, \cite{add12}. Molecular generation has emerged as a critical task in drug discovery pipelines, where creating new molecules with desired properties is a key research objective \cite{add10}. However, exploring the vast chemical space, which is estimated to contain over $10^{60}$ molecules, requires significant time and resources \cite{add8}. Currently, more than $10^8$ small molecules have been synthesized, representing only a fraction of this immense space \cite{intro4}. Despite this limited exploration, it has already made substantial contributions to human progress. This indicates that efficiently navigating the uncharted chemical space holds immense potential value.

Traditional virtual screening methods have accelerated the exploration of chemical space \cite{intro5}, \cite{intro6}. However, these approaches are limited by their reliance on enumerated molecular sets, resulting in the exploration of restricted chemical space and dependence on human knowledge \cite{intro8}. More recently, advances in deep learning-driven molecular generation have opened new avenues for efficiently exploring vast chemical spaces \cite{rev1}, \cite{rev2}. These models can probabilistically sample molecules that have not been observed in existing chemical spaces and generate potential drug candidates with desired properties \cite{intro11}, \cite{intro12}.

Simplified Molecular Input Line Entry System (SMILES) \cite{smiles_ref}-based molecular generative models have undergone extensive experimental validation and have been widely adopted \cite{add11}, \cite{add9}. However, these models require large amounts of training data, and the generated sequences must adhere to specific syntax rules and maintain chemical validity. This implies that molecular generative models can produce invalid SMILES sequences, thereby limiting the exploration of the chemical space learned by the model. Specifically, the generation of such invalid molecules hinders the proper decoding of the intended molecules by the model, which in turn degrades the diversity and quality of the generated molecules. Additionally, the range of candidate molecules with target properties is reduced, making it difficult to precisely control desired properties \cite{intro17}. This challenge is further exacerbated in cases with limited training data, such as de novo drug design, ultimately reducing opportunities for drug discovery \cite{intro19}. These challenges impose significant constraints on the practical application of molecular generative models and impede further advancements in this field.

Meanwhile, significant efforts have been made to address these challenges. To enhance the generation of valid molecules, alternative SMILES representations, such as DeepSMILES \cite{intro20} and Self-Referencing Embedded Strings (SELFIES) \cite{intro21}, have been introduced. Additionally, methods utilizing context-free grammar and attribute grammar have been attempted, though they have not been widely adopted. Recent studies have reported that such attempts limit the learning of diverse latent spaces where molecules are embedded, thereby reducing the diversity, novelty, and other evaluation metrics of the generated molecules \cite{intro24}. Graph-based models can facilitate the generation of valid molecular outputs, but their high computational cost and slow generation speed make their efficient application challenging \cite{intro17}. Some studies have drawn inspiration from translation models to attempt post-hoc corrections, making efforts to transform invalid molecules into valid ones. However, these approaches remain difficult to generalize across different generative models \cite{intro27}.

To address these limitations, we propose ChemFixer—a model that corrects invalid molecules into valid ones. ChemFixer is a framework redefined for this task by integrating masked techniques\cite{MLM1} and transformer architecture \cite{transformer}, inspired by the pre-training process of BART \cite{BART}, making it effective for this task. To enhance transferability and generalization across models, ChemFixer is pre-trained using masked techniques and fine-tuned on a large-scale dataset of valid/invalid molecular pairs that we constructed based on the widely used MOSES dataset \cite{intro28} for molecular generative models.

Transformer-based VAE frameworks have demonstrated performance across various domains of generative modeling \cite{TVAE}, \cite{TCVAE}. Inspired by this framework, we tailored it to our task and data characteristics by modifying it to effectively embed discrete molecular sequences and properties into a continuous latent space. This tailored framework was employed as the base molecular generative model. During the process of training the model to reconstruct molecules, we collected invalid molecules generated at each epoch and paired them with valid labels to build a new dataset of valid/invalid molecule pairs. This dataset serves as the foundation for fine-tuning ChemFixer.

ChemFixer significantly improved the validity of molecules generated by the base molecular generative model and demonstrated competitive performance on various benchmark metrics. Additionally, ChemFixer demonstrated the ability to preserve target molecules by correcting invalid molecules that could be lost during decoding. This indicates that ChemFixer can effectively expand the accessible region of chemical space and, furthermore, contribute to the exploration of potential molecular candidates in drug design, where specific biological properties have been enhanced or suppressed. These findings highlight the expanded possibilities and visions for drug discovery using molecular generative models.

To evaluate ChemFixer’s applicability to other models, we applied it—without additional training—to MolGCT \cite{intro11}, MolGPT \cite{intro12}, and several benchmark models supported by MOSES \cite{intro28}, including VAE \cite{vae}, \cite{intro9}, AAE \cite{aae1}, \cite{vae1}, and CharRNN \cite{rnn1}. ChemFixer improved validity while maintaining competitive performance across key evaluation metrics in all models. Additionally, within MolGCT \cite{intro11}, which supports target molecule generation, ChemFixer uncovered promising molecular candidates lost during decoding. This demonstrated ChemFixer's effective applicability across diverse models.

Notably, ChemFixer demonstrated remarkable performance in the drug–target interaction (DTI) prediction task. When the pre-trained ChemFixer was fine-tuned on the DTI prediction model Co-VAE \cite{intro18}, it significantly improved the validity of generated ligands by more than 30\%, successfully uncovering novel ligands. Consequently, it discovered novel ligand-protein pairs with competitive potential. Given that the KIBA dataset \cite{KIBA1}, \cite{KIBA2} used in this task consists of
approximately 2100 drugs and 230 targets, ChemFixer demonstrated exceptional performance even with limited data.

In the ablation study, we confirmed that ChemFixer, by internalizing SMILES grammar and error patterns through masked pre-training on a large-scale corpus, operates effectively not only within the same domain but also in chemically distinct or data-scarce environments.

Simultaneously, we conducted experiments on each model to evaluate whether ChemFixer distorts the data distribution when correcting invalid molecules into valid ones. The results showed competitive performance in benchmarks such as Fréchet ChemNet Distance (FCD) \cite{fcd}, which reflects chemical and biological properties, and Similarity to a Nearest Neighbor (SNN) \cite{intro28}, which is based on Tanimoto similarity \cite{tanimoto} to assess drug similarity. This confirms that ChemFixer preserves data distribution and properties.

In summary, our proposed ChemFixer offers the following main contributions:
\begin{itemize}
\item To the best of our knowledge, ChemFixer is the first framework designed to correct invalid molecules into valid ones while preserving the original data distribution, as well as the chemical and biological properties of the generated molecules, to improve the validity of molecular generation models.
\item We proposed a method to construct a new large-scale dataset consisting of valid/invalid molecular pairs and to train on it effectively.
\item ChemFixer demonstrated its effectiveness not only across various molecular generation models within the same domain but also in a DTI task. It maintained high validity even in data-limited settings and recovered promising candidates that were previously ungenerated due to invalidity, indicating its role as a practical and extensible framework for deep learning–based drug discovery.
\end{itemize}

\section{RELATED WORKS}
\label{sec:related works}
\subsection{Molecular Generative Models}
In recent years, numerous architectures have been developed for efficient molecular generation. Representative examples include molecular generative models based on RNNs \cite{rnn1}, GANs \cite{gan}, VAEs \cite{vae}, and Transformers \cite{transformer}. In particular, since Gómez-Bombarelli et al. \cite{intro9} demonstrated the potential of the VAE model, VAE-based molecular generative models have been integrated into various architectures and continue to serve as cornerstones in molecular generation research \cite{trvae1}.

VAE models primarily consist of two neural networks: an encoder and a decoder. During training, the encoder maps molecules onto a continuous latent space, and the decoder learns to reconstruct molecules from this latent space. Notably, well-trained latent spaces enable the generation of molecules with desired properties. However, generating target molecules by conditioning on specific properties in the latent space is challenging due to the inherently discrete nature of chemical space and the high-dimensional manifold structure. In particular, considering that generated molecules must satisfy chemical syntax rules, controlling latent vectors becomes an even more complex problem \cite{intro21}. Posterior collapse, often occurring in the VAE decoder during long sequence generation, can further complicate this process \cite{collapse}. These difficulties reduce the potential utilization of the learned chemical space.

Transformers have achieved significant success in the field of natural language processing (NLP), becoming one of the most widely used architectures in deep learning. These achievements, which enable the efficient processing of long sequences, encouraged the utilization of transformers for SMILES strings and contributed to their application in molecular generative models. Bagal et al. \cite{intro12} proposed a transformer-based SMILES molecular generative model capable of generating molecules that satisfy specific properties. Dobberstein et al. \cite{intro16} designed a model that increased the number of transformer decoder blocks to generate organic molecules with multiple properties. Hong et al. \cite{mol_tr} proposed a model that replaced the transformer decoder with feed-forward layers to address the overfitting problem, and this model was attempted to generate drug candidates for COVID-19.

Recent studies have investigated the integration of Transformers and VAE models for molecular generation. Kim et al. \cite{intro11} presented a model combining the attention mechanism of Transformers with Conditional-VAE, facilitating the effective learning of molecular properties. Dollar et al. \cite{collapse} developed a model that incorporates self-attention into $\beta$-VAE to capture complex molecular syntax rules.

\begin{figure}[t]
\begin{center}
\begin{tabular}{c @{\hspace{0.75em}} c @{\hspace{0.75em}} c}
\includegraphics[height=2cm]{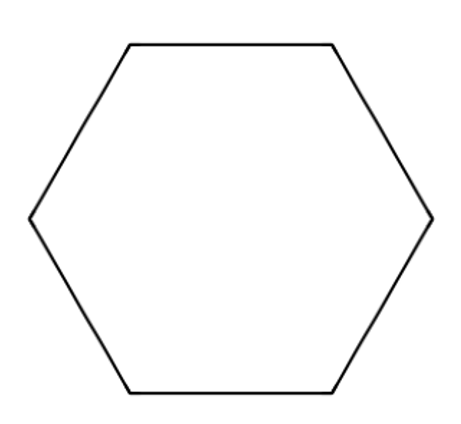} &
\includegraphics[height=2cm]{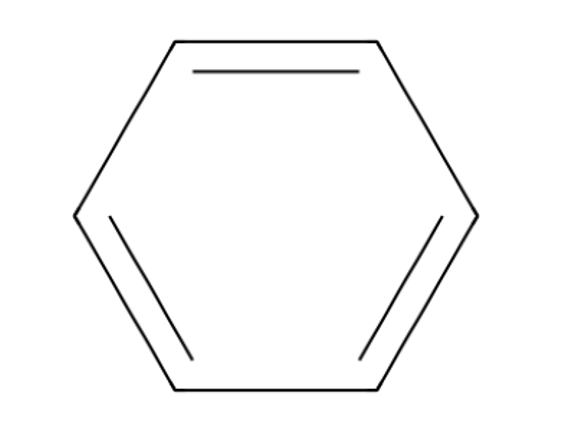} &
\includegraphics[height=2cm]{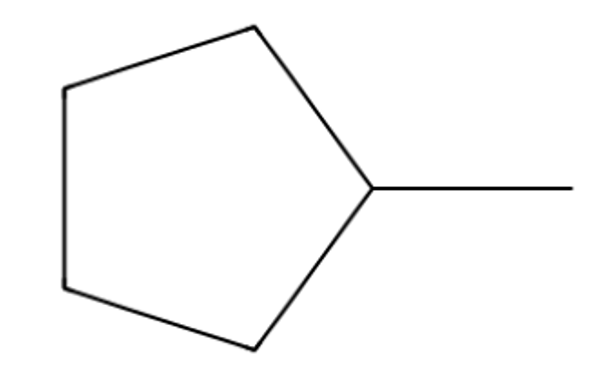} \\

\textcolor{black}{\small{C$_6$H$_{12}$}} & 
\textcolor{black}{\small{C$_6$H$_{6}$}} & 
\textcolor{black}{\small{C$_6$H$_{12}$}} \\

\textcolor{black}{\small{C1CCCCC1}} & 
\textcolor{black}{\small{c1ccccc1c}} & 
\textcolor{black}{\small{C1(C)CCCC1}} \\

\textcolor{black}{\small{(a) Cyclohexane}} & 
\textcolor{black}{\small{(b) Benzene}} & 
\textcolor{black}{\small{(c) 1-Methylcyclopentane}} \\

\end{tabular}
\end{center}
\caption{In SMILES strings, uppercase and lowercase letters, as well as branch symbols, differentiate the molecules. Uppercase C represents a non-aromatic carbon, while lowercase c represents an aromatic carbon, differentiating (a) and (b). The branch symbol indicates the position of the side chain, differentiating (a) and (c).}
\label{fig1}
\end{figure}

\subsection{Efforts to Improve the Validity of Generated Molecules}
\subsubsection{Alternative SMILES Representations}
{O’Boyle et al. \cite{intro20} proposed DeepSMILES, an alternative representation designed to address the invalidity issues arising from rings and branches during molecule decoding. Although this representation achieved better validity compared to the original SMILES representation, studies have reported that the addition of unnecessary syntax negatively affects model training and degrades performance \cite{non-smiles}, \cite{deepsmiles_low}. Krenn et al. \cite{intro21} proposed the SELFIES representation, which redefines the structure of atoms and bonds and introduces new index symbols for rings, aromaticity, and branches. This approach ensures the generation of only fully valid molecules. Skinnider \cite{intro24} recently reported that the SELFIES representation, compared with the SMILES representation, limits the learning of diverse latent spaces for molecular generation. This leads to biases in the learned chemical space and a decrease in generalization performance, as demonstrated by specific experimental results. Other representations, such as GroupSELFIES \cite{groupselfies} and GenSMILES \cite{gensmiles}, have also been proposed but have not been widely adopted for similar reasons.

\subsubsection{Post-Hoc Corrections}
{Several post-hoc approaches have been proposed to correct invalid molecules. Bilsland et al. \cite{posthoc1} applied a spelling error correction strategy originally proposed by Li et al. \cite{posthoc1_ref} to handle molecular syntax errors. Zheng et al. \cite{intro27} introduced a transformer-based model for a similar purpose. While both methods showed partial effectiveness as post-processing modules, their task-specific design limited their scalability. Recently, Zheng et al. \cite{intro17} applied techniques proposed for grammatical error correction \cite{GEC2} to manually introduce errors into a valid SMILES syntax, thereby generating an invalid syntax. Based on the resulting molecular pairs dataset, they proposed the SMILES corrector. However, since chemical syntax has unique characteristics that differ from natural language, correctors trained on data pairs with intentionally injected grammatical errors face limitations in preserving the distinct chemical space learned by the generator.}

\subsection{Data Challenges in De Novo Drug Design}
{De novo drug design in the field of deep learning applications encompasses the entire process of molecular generation, including target molecule identification, DTI, and peptide design \cite{add15}. Recent studies have demonstrated that it is possible to generate target molecules with desired properties by modulating one or more molecular conditions \cite{intro11}, \cite{intro12}. However, this process requires training data ranging from $10^{3}$ to potentially $10^{7}$. Such data dependence has been repeatedly reported to hinder further progress in de novo drug design using deep learning \cite{intro19}, \cite{add13}. Bjerrum et al. \cite{denovo4} proposed a data augmentation method that uses multiple molecular strings as inputs for the same molecule. As a result, the quality and diversity of the generated molecules could be improved, but increasing the level of data augmentation led to a decrease in model performance \cite{intro19}, \cite{denovo4}. Since this data augmentation method also requires a certain amount of data, it cannot provide a complete solution. There have also been attempts to apply the knowledge of pre-trained models to target tasks with low data through transfer learning \cite{denovo6}. However, in the case of molecular generation tasks, it is difficult to expect the same level of success as in computer vision and natural language processing, as the datasets and target molecules differ for each task \cite{denovo9}. Especially in the case of the DTI task, which is one of the key stages of de novo drug design, the relatively small amount of data used for training results in low validity of the generated molecules. This suggests that promising molecular candidates may be lost during decoding.

In other areas of de novo drug design, data insufficiency has also been a significant issue. The process of obtaining such data is not an easy task, as it involves substantial experimental and computational costs. Therefore, there is a need for more efficient methods to address these challenges.}

\begin{figure*}[!t]
    \centering
    \includegraphics[width=\textwidth]{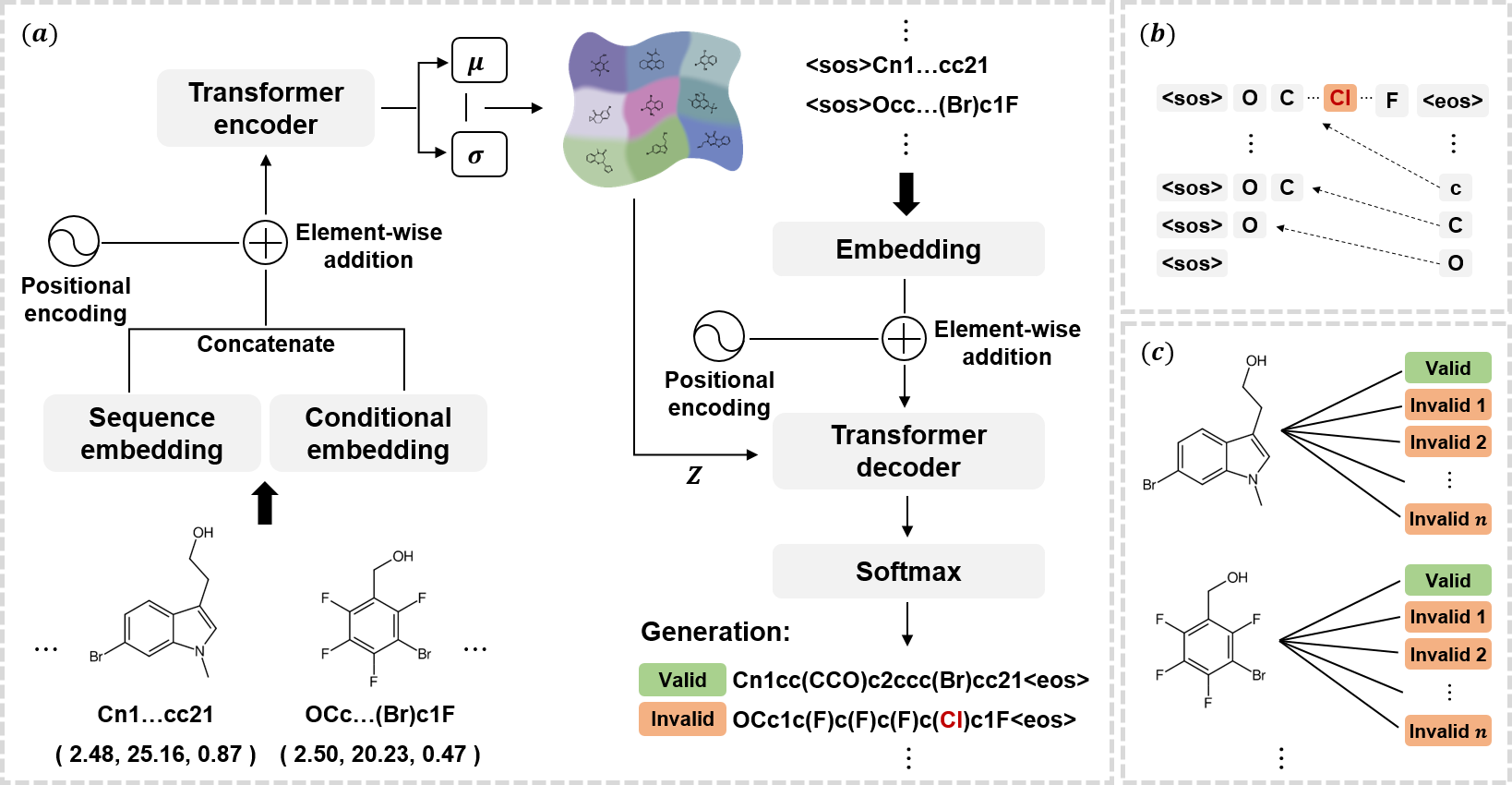} 
    \caption{Illustrations of (a) the base molecular generative model framework, (b) the process of autoregressively generating SMILES strings, and (c) valid/invalid molecular pairs generated during the training of (a).}
    \label{fig2}
\end{figure*}

\begin{figure}[!t]
\centerline{\includegraphics[width=1\columnwidth]{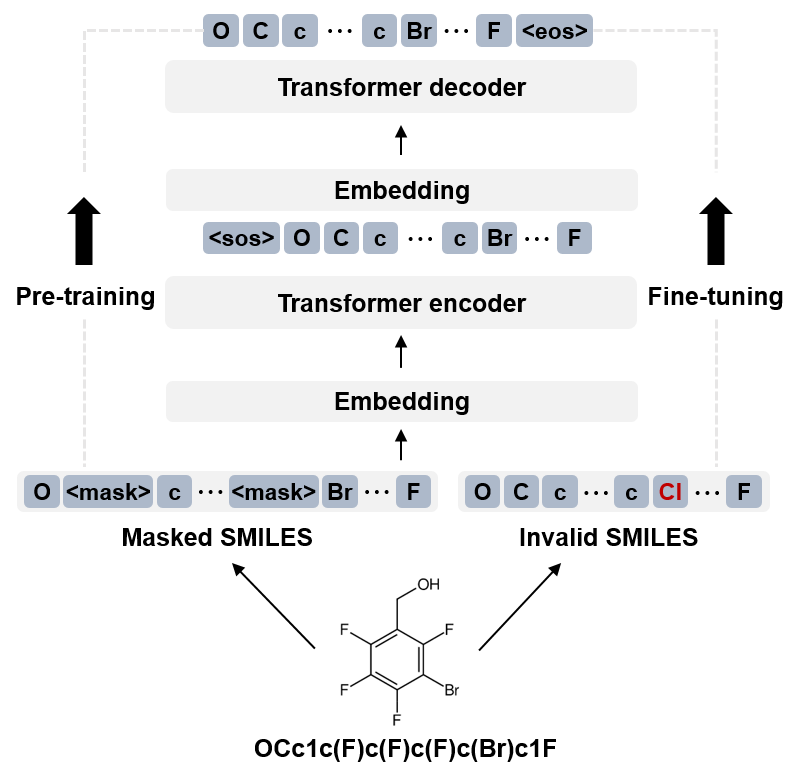}}
\caption{Illustration of ChemFixer: pre-training with a 10\% masking ratio followed by fine-tuning using valid/invalid molecular pairs.}
\label{fig3}
\end{figure}

\section{MATERIALS AND METHODS}{
\subsection{SMILES String Tokenization}
The process of tokenizing SMILES strings is necessary to embed them into a deep learning model. In this study, we adopted the rule-based (atom-wise) tokenizer from SmilesPE \cite{smilespe}. A fixed vocabulary of 29 tokens was constructed by excluding the charge symbols ($+$ and $-$), which do not appear in the MOSES dataset. The maximum sequence length was set to 80 tokens.

The token vocabulary used in this study is categorized as follows:

\vspace{0.5\baselineskip}
\hangindent=\parindent 
\hangafter=1 
{\it \textbf{13 atom tokens:}} {
\textnormal{\textless C\textgreater, \textless c\textgreater, \textless O\textgreater, \textless o\textgreater, \textless N\textgreater, \textless n\textgreater, \textless S\textgreater, \textless s\textgreater, \textless F\textgreater, \textless Cl\textgreater, \textless Br\textgreater, \textless[H]\textgreater, and \textless[nH]\textgreater}
}

\vspace{0.5\baselineskip}
\hangindent=\parindent 
\hangafter=1 
{\it \textbf{6 ring tokens:}} {
\textnormal{\textless 1\textgreater, \textless 2\textgreater, \textless 3\textgreater, \textless 4\textgreater, \textless5\textgreater, and \textless6\textgreater}
}

\vspace{0.5\baselineskip}
\hangindent=\parindent 
\hangafter=1 
{\it \textbf{3 bond tokens:}} {
\textnormal{\textless --\textgreater, \textless =\textgreater, and \textless\#\textgreater}
}

\vspace{0.5\baselineskip}
\hangindent=\parindent 
\hangafter=1 
{\it \textbf{2 branch (side chain) tokens:}} {
\textnormal{\textless (\textgreater \ and \textless )\textgreater}
}

\vspace{0.5\baselineskip}
\hangindent=\parindent 
\hangafter=1 
{\it \textbf{5 special tokens:}} {
\textnormal{\textless sos\textgreater, \textless eos\textgreater, \textless pad\textgreater, \textless unknown\textgreater, and \textless mask\textgreater}
}
\vspace{0.5\baselineskip}

In SMILES, lowercase \textnormal{\textless c\textgreater, \textless o\textgreater, \textless n\textgreater, \textless s\textgreater} represent atoms that are part of an aromatic ring, whereas uppercase \textnormal{\textless C\textgreater, \textless O\textgreater, \textless N\textgreater, \textless S\textgreater} represent non-aromatic atoms or atoms bonded to the aromatic ring but not part of it. In the case of hydrogen, it can be inferred based on chemical bonding rules and is usually omitted in SMILES strings. However, when bond information needs to be explicitly specified, it is represented as \textnormal{\textless [H]\textgreater}, and \textnormal{\textless [nH]\textgreater} represents a hydrogen atom bonded to the nitrogen atom \textnormal{\textless n\textgreater}, which is part of an aromatic ring. Ring structures are indicated by repeating the same number to mark the start and end of the ring. As the number of rings increases, numbers such as {\textless 2\textgreater}, {\textless 3\textgreater}, etc., can be used to represent additional rings. The 3 bond tokens {\textless --\textgreater}, {\textless =\textgreater}, {\textless \#\textgreater} represent single bond, double bond, and triple bond, respectively. The branch symbol indicates the position of the side chain. {\textless sos\textgreater} and {\textless eos\textgreater} represent the start and end tokens, respectively. {\textless pad\textgreater} is a token used to fill the pre-defined sequence length, while {\textless unknown\textgreater} refers to tokens not included in this set. {\textless mask\textgreater} token is a string masking token used during the training process of ChemFixer. Charged atoms and stereochemical tokens that were not included in the MOSES dataset were excluded. Fig. \ref{fig1} shows a simple example of SMILES string differentiation, and the usage of tokens is explained in \textit{B. Generation of Valid/Invalid Molecular Pairs}.
}

\subsection{Generation of Valid/Invalid Molecular Pairs}
This section describes the process of constructing a dataset for training ChemFixer. During the training of the base molecular generative model, invalid molecular data is generated and labeled with valid molecules to create a valid/invalid molecular pairs dataset.

\subsubsection{Invalid Molecular Generation Framework}
The transformer-based VAE framework was adopted as the base molecular generative model, in which SMILES strings (\(M_{s}\)) and their corresponding properties (\(M_{p}\)) were used for training to generate molecules.

Given a SMILES molecule (\(M_{s}\), \(M_{p}\)) , the SMILES string is tokenized into integers according to predefined tokens and organized as:
\begin{equation} \label{eq1}
  M_{s} \in \mathbb{N}^{b \times l}, 
  \qquad
  M_{p} \in \mathbb{R}^{b \times c},
\end{equation}
\noindent
where $b$ is the batch size, $l$ is the maximum SMILES length, and $c$ is the property dimension.
\begin{equation} \label{eq2}
\begin{aligned}
E_{s}:\,\mathbb{N} \;\to\;\mathbb{R}^{d_{\mathit{model}}}, 
\quad & E_{s}\bigl(M_{s}[b,:]\bigr) \;\in\;\mathbb{R}^{l \times d_{\mathit{model}}},\\[6pt]
E_{p}:\,\mathbb{R}\;\to\;\mathbb{R}^{d_{\mathit{model}}}, 
\quad & E_{p}\bigl(M_{p}[b,:]\bigr) \;\in\;\mathbb{R}^{c \times d_{\mathit{model}}}.
\end{aligned}
\end{equation}
\noindent

After embedding the sequences and properties, the embeddings are concatenated and processed with positional encoding (\(\displaystyle P \in \mathbb{R}^{(l + c)\times d_{\mathit{model}}}\)) as follows:
\begin{equation} \label{eq3}
\Bigl(
  E_{s}\bigl(M_{s}[b,:]\bigr)
  \;\oplus\;
  E_{p}\bigl(M_{p}[b,:]\bigr)
\Bigr)
+ P
\;\in\;\mathbb{R}^{(\,l + c\,)\times d_{\mathit{model}}}.
\end{equation}
\noindent

Let \eqref{eq3} be \( M_{\mathit{in}}^{\mathit{en}}[b] \), The operations in the transformer encoder are as follows:
\begin{equation} \label{eq4}
\begin{aligned}
Q = M_{\mathit{in}}^{\mathit{en}}[b]\,W^{Q},
\quad & K = M_{\mathit{in}}^{\mathit{en}}[b]\,W^{K},\\[6pt]
V = M_{\mathit{in}}^{\mathit{en}}[b]\,W^{V}, 
\quad & W^{Q},\,W^{K},\,W^{V} \in \mathbb{R}^{d_{\mathit{model}} \times d_{\mathit{head}}},
\end{aligned}
\end{equation}
\begin{equation} \label{eq5}
\begin{gathered}
\alpha_{t,t'} = \text{softmax}_{t'}\!\Bigl(\frac{Q[t,:] \cdot K[t',:]}{\sqrt{d_{\mathit{head}}}}\Bigr),\\[6pt]
\text{Attention}(Q, K, V)[t,:] = \sum_{t'} \alpha_{t,t'}\,V[t',:],
\end{gathered}
\end{equation}
\noindent
where $t, t' \in \{1, \ldots, (l + c)\}.$ In the encoder, operation \eqref{eq4} linearly transforms the trainable parameters  \(\,(W^Q, W^K, W^V)\) with 
\(\,M_{\mathit{in}}^{\mathit{en}}[b]\), thereby performing multi-head self-attention. Through scaled dot-product attention, the similarity \(\alpha_{t,t'}\) and 
\(\mathrm{Attention}(Q,K,V)[t,:]\) are computed among sequence tokens and property tokens, forming a context vector that integrates both the sequence and property information of the molecule. The multiple-head operation applies unique parameters 
\((W^Q_i, W^K_i, W^V_i)\) in parallel to repeat the same procedure, and after which, each head’s output is combined via concatenation and a linear transformation. Likewise, the decoder also employs multi-head attention in both masked self-attention and cross-attention, following the same parallel heads and concatenation procedure.

Let the output of the encoder be \(M_{\mathit{out}}^{\mathit{en}}[b]\), and the latent space is obtained as follows:
\begin{equation} \label{eq6}
\begin{gathered}
E_{\mathit{mol}}^{\mu}, \ E_{\mathit{mol}}^{\mathit{\sigma}} : 
\mathbb{R}^{(c + l)\times d_{\mathit{model}}} 
\longrightarrow 
\mathbb{R}^{(c + l)\times d_{z}}, \\
\mu_{\mathit{mol}} = 
E_{\mathit{mol}}^{\mu}\bigl(M_{\mathit{out}}^{\mathit{en}}[b]\bigr), \
\sigma_{\mathit{mol}} = 
E_{\mathit{mol}}^{\mathit{\sigma}}\bigl(M_{\mathit{out}}^{\mathit{en}}[b]\bigr),
\end{gathered}
\end{equation}\begin{equation} \label{eq7}
Z = \mu_{\mathit{mol}} \;+\; \sigma_{\mathit{mol}} \;\epsilon,
\
\epsilon \,\sim\, \mathcal{N}\!\bigl(\mathbf{0}, \mathbf{I}\bigr).
\end{equation}

\indent
For teacher forcing in the decoder, sequences with added special tokens and their corresponding properties are embedded individually, similar to \eqref{eq2}. The subsequent processes of concatenation and positional encoding are the same as in \eqref{eq3} and are computed as follows:
\begin{equation} \label{eq8}
\begin{aligned}
E_{s'} : \mathbb{N} \to \mathbb{R}^{d_{\mathit{model}}}, \qquad
E_{p'} : \mathbb{R} \to \mathbb{R}^{d_{\mathit{model}}}, \qquad \\
\Bigl(
  E_{s'}\bigl(M_{s'}[b,:]\bigr)
  \kern+0.1em\oplus\kern+0.1em
  E_{p'}\bigl(M_{p'}[b,:]\bigr)
\Bigr)\kern-0.1em+\kern-0.1em P
\in\mathbb{R}^{(l' + c)\times d_{\mathit{model}}},
\end{aligned}
\end{equation}
where \(l'\) is the maximum length. Let \eqref{eq8} \(\,M_{\mathit{self}}^{\mathit{de}}[b]\). 
In the transformer decoder, the masked self-attention Query ($Q$), Key ($K$), and Value ($V$) are computed through operations with \(\,M_{\mathit{self}}^{\mathit{de}}[b]\),  in the same manner as \eqref{eq4}. This process blocks future
tokens to enable autoregressive generation, and the masking
is performed as follows:
\begin{equation} \label{eq9}
\mathrm{Score}_{\text{self}}(s,s')
=
\begin{cases}
\dfrac{Q[s,:] \cdot K[s',:]}{\sqrt{d_{\mathit{head}}}}, & \text{if } s' \le s,\\[6pt]
-\infty, & \text{if } s' > s,
\end{cases}
\end{equation}
where $\mathit{s}, \mathit{s}' \in \{1,\dots,(\ell' + c)\}.$ 
Applying $\mathrm{Score}_{\text{self}}(s,s')$ to the softmax makes any position with $s' > s$ become zero, 
so each token $\mathit{s}$ cannot refer to tokens beyond $\mathit{s}$. 
Consequently, the decoder can generate a sequence in an autoregressive manner. Let the output of this self-attention be denoted as \(\,M_{\mathit{cross}}^{\mathit{de}}[b]\).
Following \eqref{eq4}, the cross-attention $Q$ is computed from 
\(\,M_{\mathit{cross}}^{\mathit{de}}[b]\), thereby incorporating 
the updated context, while $K$ and $V$ are computed 
from $Z$ in \eqref{eq7}, allowing the model to reference both the molecular sequence and properties.

After the cross-attention and subsequent layers, 
the decoder produces its final hidden state 
\(M_{\mathit{out}}^{\mathit{de}}[b] \in \mathbb{R}^{(l' + c)\times d_{\mathit{model}}}\), and the probability distribution over the token $v$ is as follows: 
\begin{equation} \label{eq10}
\mathit{o}_s =
\mathit{W}_{\mathit{out}}
\;\mathit{M}_{\mathit{de}}^{\mathit{out}}[b][s], \;\;
p\bigl(y_s = v\bigr)
= \mathrm{softmax}\bigl(\mathit{o}_s\bigr)[v],
\end{equation}
\noindent
where $W_{\mathit{out}}\in\mathbb{R}^{\,d_{\mathit{model}}\times\mathit{token\_size}}$ 
serves as the final linear transformation parameter, 
and $\mathit{token\_size}$ denotes the size of the token vocabulary. Consequently, as shown in Fig. \ref{fig2}(b), the decoder can generate the molecule one token at a time in an autoregressive manner.

We train the model by jointly considering the cross-entropy loss 
$L_{\mathit{mol}}$ between the generated molecule and its label, 
the KL-divergence loss $L_{\mathit{KL}}$ for the VAE, 
and the property prediction loss $L_{\mathit{prop}}$. 
The mean squared error (MSE) is employed when predicting $c$ properties. The final training loss $\mathcal{L}$ is defined as follows:
\begin{equation}\label{eq11}
\mathcal{L}
\;=\;
L_{\mathit{mol}}
\;+\;
\beta\,L_{\mathit{KL}}
\;+\;
L_{\mathit{prop}}.
\end{equation}

\subsubsection{Valid/Invalid Molecular Pairs}
In the base molecular generative model, at each epoch \( e_i \in \{e_1, \ldots, e_n\} \),  an input pair (\(M_{s}, M_{p})\) is provided, and the model is trained to reconstruct the same SMILES string as follows:
\begin{equation} \label{eq12}
\hat{M} = \mathit{model}^{\mathit{rec}} (M_s, M_p).
\end{equation}

The validity of the reconstructed molecule $\hat{M}$ is evaluated using an open-source chemical parsing library \cite{rdkit}. If $\hat{M}$ is determined to be invalid, it is stored along with its input pair (\(M_{s}, M_{p})\) and used as a training sample. For each epoch \(e_i\), such invalid outputs are collected and defined as follows:
\begin{equation} \label{eq13}
\begin{split}
I(e_i) = \{(\hat{M}, \mathit{label}) \mid \hat{M} = \text{invalid}\}, \\
D_{\text{label/invalid}} = \bigcup_{i=1}^{n} I(e_i). \qquad \qquad
\end{split}
\end{equation}

As shown in Fig. \ref{fig2}(c), invalid data are collected for each molecule at every epoch. The model can generate multiple invalid molecules both within a single epoch and across different epochs.

However, previous studies \cite{molgan} have reported that generated molecules tend to be less reliable in the early stages of training. Therefore, in this study, we selectively collected invalid samples only from epochs where the distribution of generated molecules appeared reasonably similar to that of real molecules. Specifically, in each epoch, we sampled 3,000 valid molecules and evaluated not only the validity of generated samples but also their distributional similarity to real molecules using Fréchet ChemNet Distance (FCD) \cite{fcd}, as well as the distribution of logP, one of the key molecular properties.

\subsection{ChemFixer}
As shown in Fig. \ref{fig3}, ChemFixer is first pre-trained on masked SMILES strings, then fine-tuned on the previously constructed $D_{\text{label/invalid}}$ dataset, and trained in a manner similar to the procedures described in \eqref{eq1}–\eqref{eq11} of the base molecular generative model.
\subsubsection{Pre-training Process}
During the process of embedding a SMILES string into the model \eqref{eq1}, \eqref{eq2}, each token $t_{i}$ of the input SMILES string $S$ is masked with a probability of 10\%, following a Bernoulli distribution $M_i \sim \mathit{Bernoulli}(0.10)$, as follows:
\begin{equation} \label{eq14}
\begin{split}
S &= (t_1, t_2, \dots, t_l) \quad \rightarrow \quad S' = (t'_1, t'_2, \dots, t'_l), \\
&\quad\qquad t'_i =
\begin{cases}
\langle \text{mask} \rangle, & \text{if } M_i = 1, \\
\quad t_i,           & \text{if } M_i = 0,
\end{cases}
\end{split}
\end{equation}
where $l$ is the maximum SMILES length. Then, ChemFixer $\mathit{f}_{\theta_{\mathit{pre}}}(\cdot)$ is trained in an autoregressive manner via a transformer encoder \eqref{eq3}-\eqref{eq5} and decoder \eqref{eq8}-\eqref{eq10} so that the input $S'$ is reconstructed as $S$,
\begin{equation}\label{eq15}
\hat{S} = \mathit{f}_{\theta_{\mathit{pre}}}(S'), \qquad \hat{S} = (\hat{t_1}, \hat{t_2}, \dots, \hat{t_l}),
\end{equation}
For each token position $i$, the cross-entropy loss $\ell(\hat{t}_i, t_i)$ is computed between the predicted token $\hat{t}_i$ and the label token $t_i$, and the loss function is as follows:
\begin{equation}\label{eq16}
L_{\mathit{pre}} 
= \sum_{i=1}^{l} \ell(\hat{t}_i, t_i).
\end{equation}

Since ChemFixer predicts one token at each position from a fixed 29-token SMILES vocabulary, the task is inherently categorical. Accordingly, cross-entropy loss—rather than regression losses such as mean squared error (MSE) or mean absolute error (MAE)—provides the most principled objective for minimizing the divergence between the predicted and ground-truth distributions. This approach has also been widely adopted in prior works such as ChemBERTa \cite{CE1} and X-MOL \cite{CE2}.

\subsubsection{Fine-tuning Process}
From the $D_{\text{label/invalid}}$ dataset \eqref{eq13}, invalid SMILES strings $I$ are input into ChemFixer $\mathit{f}_{\theta_{\mathit{fine}}}(\cdot)$ and fine-tuned in the same manner as pre-training, in order to reconstruct the label $V$:
\begin{equation} \label{eq17}
\begin{split}
V =& (v_1, v_2, \dots, v_l), \\
\hat{V} = \mathit{f}_{\theta_{\mathit{fine}}}(I),& \qquad 
\hat{V} = (\hat{v}_1, \hat{v}_2, \dots, \hat{v}_l), \\
L_{\mathit{fine}} &= \sum_{i=1}^{l} \ell(\hat{v}_i, v_i).
\end{split}
\end{equation}

For each token position \( i \), the cross-entropy loss \( \ell(\hat{v}_i, v_i) \) is computed between the predicted token \( \hat{v}_i \) and the label token \( {v}_i \), and the overall loss function is defined as \( L_{\mathit{fine}} \).

\subsection{Model Configuration}
Both the encoder and decoder adopt a standard Transformer \cite{transformer} architecture consisting of 6 layers, each containing 8 self-attention heads. Each attention head is 64-dimensional, resulting in a total embedding dimension of 512, while the position-wise feed-forward sub-layers use a hidden size of 2 048. With a batch size $B$ and a maximum sequence length of 80, the encoder and decoder inputs have a shape of ($B$, 80, 512) after token and positional embedding. A dropout rate of 0.25 is applied to all projection and attention operations. Positional encoding is separately added to both the encoder and decoder embeddings to enable the model to learn token order information, thereby preserving the structural consistency of SMILES sequences. The variational bottleneck is parameterised by a 128-dimensional latent vector $z$. ChemFixer is trained solely on SMILES sequences without utilizing any molecular property information. In contrast, the base molecular generative model incorporates three molecular properties—logP, tPSA, and QED—by embedding them into the input sequence and reusing them during the decoder’s cross-attention phase to guide molecule generation. Both models are trained using the Adam optimizer \cite{adam} with cosine learning rate scheduling \cite{sgdr}, and gradient clipping is applied to ensure stable convergence.

\begin{figure*}[!t]
    \centering
    \includegraphics[width=\textwidth]{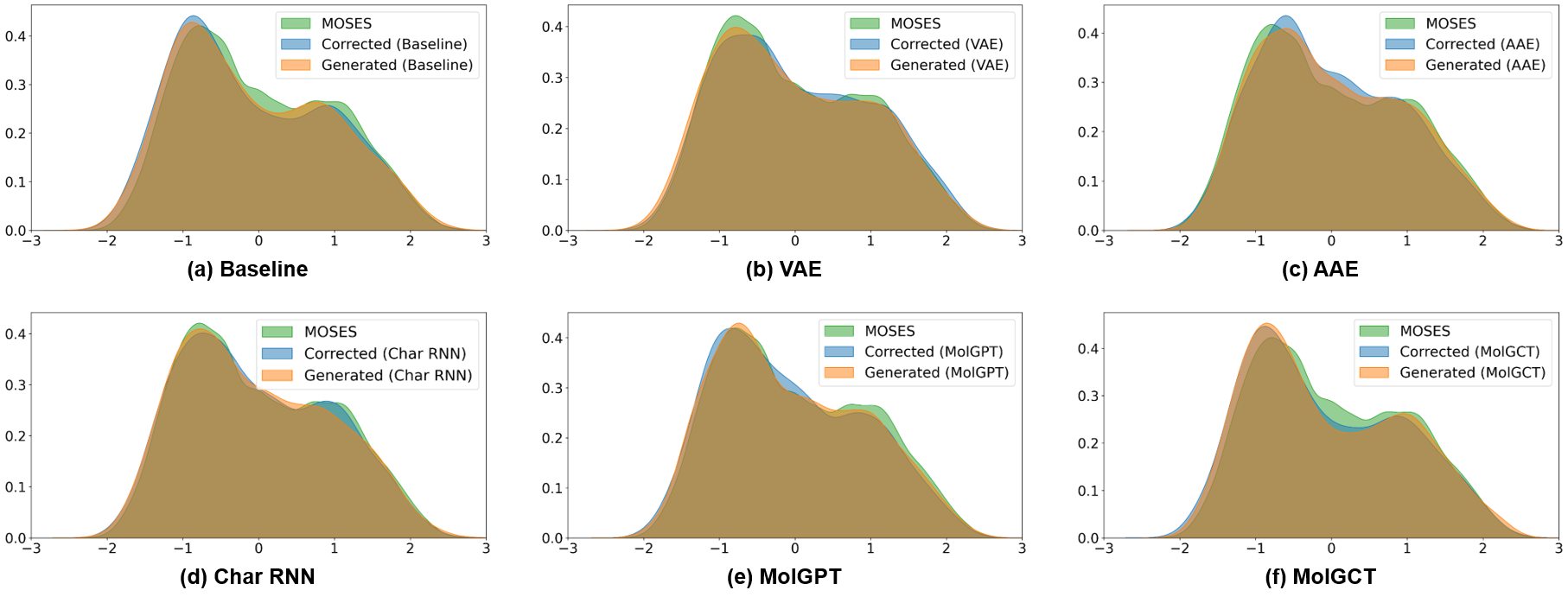} 
    \caption{Distributional comparison among 5,000 molecules each from the MOSES test data, model-generated data, and ChemFixer-corrected data. Distributions were visualized using PCA and KDE along the first principal component.}
    \label{fig4}
\end{figure*}

\begin{table}[!t]
\caption{QUANTITATIVE EVALUATION OF DATA DISTRIBUTION SIMILARITY}
\label{table1}
\centering 
\begin{tabular}{lccc}
\toprule
 & \multicolumn{3}{c}{Evaluation Metrics} \\ \cmidrule{2-4}
Methods              & FCD/Test (\textdownarrow) & SNN/Test (\textuparrow) & Scaf/TestSF (\textuparrow) \\ \midrule
Baseline             & 0.847   & 0.613   & 0.079       \\ 
+ ChemFixer (Ours)   & 0.912   & 0.602   & 0.087       \\ \midrule
VAE \cite{vae}, \cite{intro9}          & 0.295   & 0.619   & 0.050       \\ 
+ ChemFixer (Ours)   & 0.316   & 0.602   & 0.061       \\ \midrule
AAE \cite{aae1}, \cite{vae1}          & 0.821   & 0.610   & 0.071       \\ 
+ ChemFixer (Ours)   & 0.860   & 0.601   & 0.078       \\ \midrule
CharRNN \cite{rnn1}      & 0.266   & 0.607   & 0.101       \\ 
+ ChemFixer (Ours)   & 0.299   & 0.599   & 0.107       \\ \midrule
MolGPT \cite{intro12}       & 0.607   & 0.618   & 0.083       \\ 
+ ChemFixer (Ours)   & 0.638   & 0.607   & 0.089       \\ \midrule
MolGCT \cite{intro11}       & 0.762   & 0.620   & 0.086       \\ 
+ ChemFixer (Ours)   & 0.806   & 0.604   & 0.094       \\ \bottomrule
\multicolumn{4}{l}{\textuparrow: Higher is better, \textdownarrow: Lower is better} \\
\end{tabular}
\end{table}

\section{EXPERIMENTS AND RESULTS}
\subsection{Datasets and Benchmarks}
In this study, we trained and evaluated both the baseline molecular generative model and ChemFixer using the MOSES benchmark dataset \cite{intro28}.
Furthermore, to evaluate its generalizability, ChemFixer was directly applied—without any additional fine-tuning—to various other molecular generative models trained on the same dataset, including VAE \cite{vae}, \cite{intro9}, AAE \cite{aae1}, \cite{vae1}, CharRNN \cite{rnn1}, MolGPT \cite{intro12}, and MolGCT \cite{intro11}.

The MOSES dataset is derived from the ZINC database \cite{zinc} and comprises lead-like molecules considered promising in the early stages of drug discovery. More specifically, these molecules have a molecular weight of 250–350 Da, at most 7 rotatable bonds, and do not include any atoms other than C, N, S, O, F, Cl, Br, H, or charged atoms, among other constraints. MOSES is divided into three non-overlapping sets, consisting of 1.6 million molecules for training, 176,000 for testing (Test), and 176,000 for scaffold testing (TestSF). Notably, the scaffold test set is composed exclusively of molecules whose scaffolds do not appear in either of the other two sets, making it possible to assess the ability to generate entirely novel scaffolds. The uniqueness\text{@}10K (Unique\text{@}10K) metric is defined as the ratio of unique valid molecules to the total number of valid molecules among 10,000 generated samples, computed after RDKit validation \cite{rdkit}.

Additionally, we validated the performance of ChemFixer on Co-VAE \cite{intro18}, a DTI model that is not limited to molecular generation. This model uses the KIBA dataset \cite{KIBA1}, \cite{KIBA2}, which deals with binding affinities between proteins and ligands. Co-VAE excludes drug–target pairs with affinities below 10, as described in the original Co-VAE paper, resulting in a smaller dataset of 2,111 ligands and 229 proteins. We used the RDKit toolkit \cite{rdkit} to validate SMILES strings through a two-step process involving syntax and chemical-rule checks, and to perform molecular property calculations and Bemis–Murcko scaffold \cite{murcko} extraction.

\begin{figure}[!t]
\centerline{\includegraphics[width=1\columnwidth]{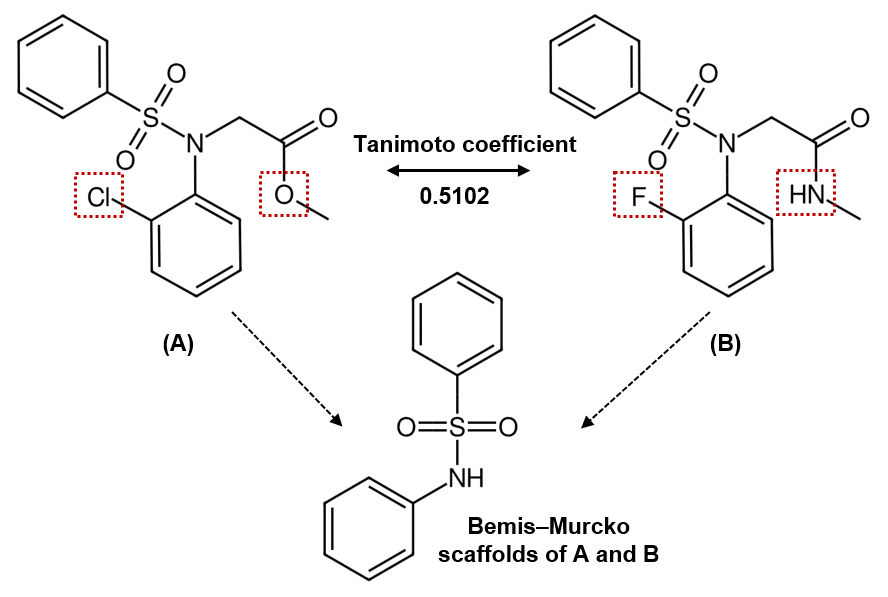}}
\caption{Example of a ChemFixer-corrected molecule with a different pharmacology but the same Bemis–Murcko scaffold.}
\label{fig5}
\end{figure}

\begin{table*}[!t]
\caption{COMPARISON OF THE VALIDITY, CORRECTION RATE, UNIQUENESS, FCD, AND SNN METRICS OF MOLECULAR GENERATION MODELS BEFORE AND AFTER CHEMFIXER CORRECTION}
\label{table2}
\centering
\begin{tabular}{@{}cclccccc@{}}
\toprule
& \multicolumn{7}{c}{Methods} \\ \cmidrule(l){2-8}
\multirow{2}{*}{\begin{tabular}[c]{@{}c@{}}Evaluation\\ Metrics\end{tabular}}
& \multicolumn{2}{c}{Baseline} & VAE \cite{vae}, \cite{intro9} & AAE \cite{aae1}, \cite{vae1}
& CharRNN \cite{rnn1} & MolGPT \cite{intro12} & MolGCT \cite{intro11} \\ \cmidrule(l){2-8}
& \multicolumn{2}{c}{+ ChemFixer (Ours)} & + ChemFixer (Ours) & + ChemFixer (Ours)
& + ChemFixer (Ours) & + ChemFixer (Ours) & + ChemFixer (Ours) \\ \midrule
\multirow{2}{*}{Validity}
& \multicolumn{2}{c}{0.9834} & 0.9713 & 0.9270 & 0.9702 & 0.9929 & 0.9901 \\
& \multicolumn{2}{c}{0.9911} & 0.9883 & 0.9793 & 0.9881 & 0.9971 & 0.9964 \\
Correction rate
& \multicolumn{2}{c}{(46.4\%)} & (59.2\%) & (71.6\%) & (60.1\%) & (59.1\%) & (63.6\%) \\ \midrule
\multirow{2}{*}{Unique@10K}
& \multicolumn{2}{c}{0.9935} & 0.9993 & 0.9960 & 0.9993 & 0.9993 & 0.9948 \\
& \multicolumn{2}{c}{0.9937} & 0.9994 & 0.9980 & 0.9995 & 0.9993 & 0.9950 \\ \midrule
\multirow{2}{*}{FCD / Test}
& \multicolumn{2}{c}{0.498 $\pm$ 0.03} & 0.099 $\pm$ 0.01 & 0.556 $\pm$ 0.20 & 0.073 $\pm$ 0.03 & 0.067 $\pm$ 0.00 & 0.402 $\pm$ 0.00 \\
& \multicolumn{2}{c}{0.513 $\pm$ 0.01} & 0.108 $\pm$ 0.01 & 0.598 $\pm$ 0.03 & 0.082 $\pm$ 0.02 & 0.068 $\pm$ 0.01 & 0.407 $\pm$ 0.02 \\ \midrule
\multirow{2}{*}{SNN / Test}
& \multicolumn{2}{c}{0.614 $\pm$ 0.02} & 0.626 $\pm$ 0.00 & 0.608 $\pm$ 0.02 & 0.601 $\pm$ 0.02 & 0.623 $\pm$ 0.02 & 0.618 $\pm$ 0.00 \\
& \multicolumn{2}{c}{0.601 $\pm$ 0.01} & 0.603 $\pm$ 0.01 & 0.595 $\pm$ 0.01 & 0.593 $\pm$ 0.02 & 0.622 $\pm$ 0.01 & 0.616 $\pm$ 0.01 \\ \bottomrule
\multicolumn{8}{l}{\footnotesize Mean ± SD (3 runs) is reported for FCD and SNN. For all other metrics, SD is below 0.01.} \\

\end{tabular}
\end{table*}

\subsection{Analysis of Data Distribution Preservation}
Experiments were conducted to demonstrate that ChemFixer corrects invalid molecules into valid ones without distorting or biasing the data distribution generated by the original model. To evaluate the data distribution preservation performance of ChemFixer, we sampled 5,000 molecules from each of the six molecular generation models, including the baseline model, and also collected 5,000 valid molecules corrected by ChemFixer from each model through repeated sampling. Subsequently, we compared the two datasets with the MOSES test set.

First, as shown in Fig. \ref{fig4}, principal component analysis (PCA) was performed on the high-dimensional molecular features to reduce their dimensionality, and a kernel density estimation (KDE) graph was plotted along the first principal component axis to evaluate the similarity of the data distributions. Overall, the three distributions for each model in Fig. \ref{fig4} exhibit similar shapes, indicating that they reflect mutually similar data characteristics without being distorted or biased toward any particular side. In particular, the MOSES test set exhibited a more refined distribution, likely because it contains far more molecules than the sampled or corrected datasets, thus reflecting a wider range of molecular characteristics. An important point is that the distribution of the corrected data is not fixed, but differs for each model that generated the invalid molecules. This suggests that ChemFixer does not simply transform invalid molecules into arbitrary valid ones, but rather corrects them into the properly labeled molecules each model intended to generate.

\begin{figure*}[!t]
    \centering
    \includegraphics[width=\textwidth]{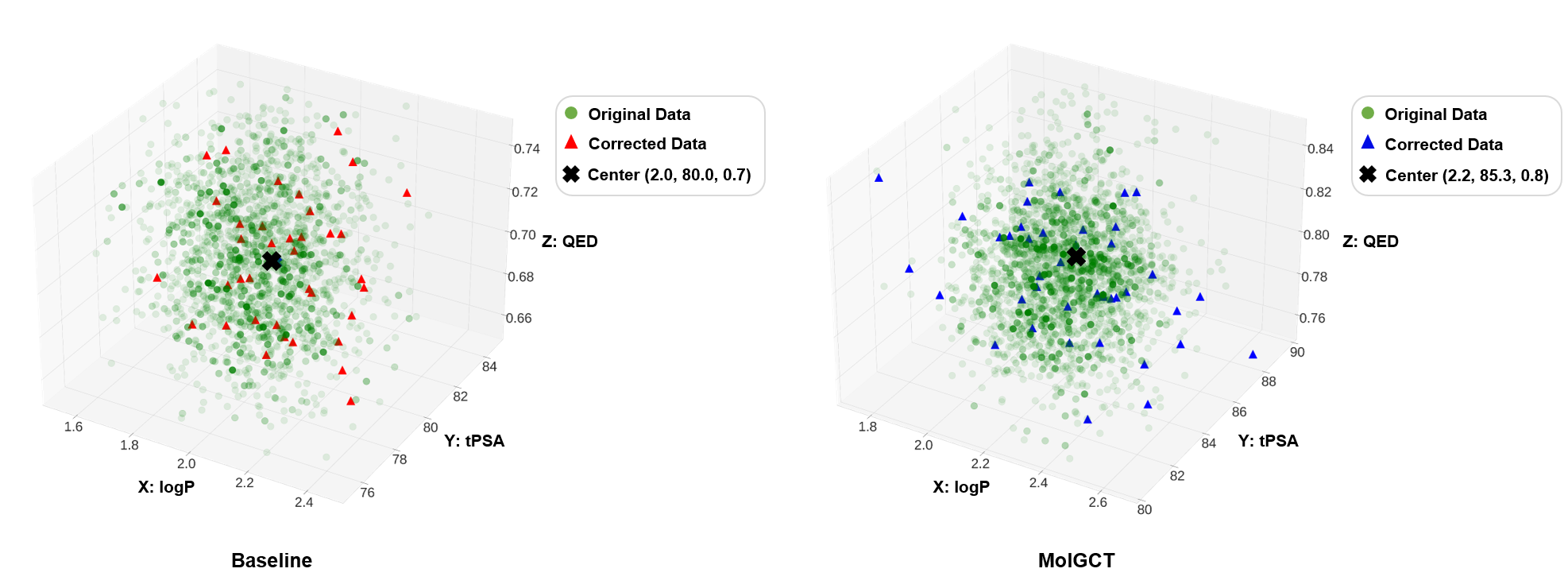} 
    \caption{Target molecule generation results: The center represents the target values, the original data shows the target molecules generated by the models, and the corrected data shows the target molecules discovered through ChemFixer, which the models could not generate.}
    \label{fig6}
\end{figure*}

Second, we conducted a quantitative evaluation using FCD and SNN, whose definitions are as follows.

\noindent
\textbf{Fréchet ChemNet Distance (FCD)} \cite{fcd} is calculated using the activations of the penultimate layer of ChemNet. Due to its advantage of providing a comprehensive assessment of whether the generated molecules are diverse and possess chemical and biological properties similar to real molecules, it is widely used as an evaluation metric for molecular generation models. FCD for two distributions ($G$, $R$) is calculated as follows:
\begin{equation} \label{eq18}
\small
    \text{FCD}(G, R) = \|\mathit{\mu}_G - \mathit{\mu}_R\|^2
    + \text{Tr}\!\Bigl(\mathit{\Sigma}_G + \mathit{\Sigma}_R
    - 2\bigl(\mathit{\Sigma}_G \mathit{\Sigma}_R\bigr)^{1/2}\Bigr),
\end{equation}
where $\mathit{\mu}$ is the mean, $\mathit{\Sigma}$ is the covariance, and $\text{Tr}(\cdot)$ is the trace operator.
\noindent
\textbf{Similarity to a Nearest Neighbor (SNN)} \cite{intro28} is an evaluation metric based on the Tanimoto similarity $T(\mathit{m_G}, \mathit{m_R})$ \cite{tanimoto} used to measure the similarity of drugs, and is calculated as follows:
\begin{equation} \label{eq19}
\text{SNN}(\mathit{G}, \mathit{R}) = \frac{1}{|\mathit{G}|} \sum_{\mathit{m_G} \in \mathit{G}} \max_{\mathit{m_R} \in \mathit{R}} T(\mathit{m_G}, \mathit{m_R}),
\end{equation}
where the Morgan fingerprints \cite{morgan} of a molecule $\mathit{m_G}$ from the generated set $G$ and its nearest neighbor molecule $\mathit{m_R}$ in the reference dataset $R$ are compared.

Table \ref{table1} presents a comparison between the data used in the previous experiment and the MOSES test dataset using FCD, SNN, and Scaf/TestSF metrics. The FCD and SNN values of ChemFixer-corrected data were similar to those of the original model-generated data. In addition, the slight increase in Scaf/TestSF—which measures the proportion of novel scaffolds—suggests that ChemFixer does not statistically compromise the chemical, biological, or drug-like properties learned by the original generative models during the correction process.

Notably, across all models, the change in FCD for the corrected data was more pronounced than the change in SNN. Since FCD accumulates squared deviations \eqref{eq18}, even localized scaffold-level changes can significantly increase the distance. In contrast, SNN, which averages the maximum Tanimoto similarity between each generated molecule and its closest neighbor in the reference set, becomes saturated at high similarity values, rendering minor structural modifications less impactful and resulting in only a marginal decrease in SNN \eqref{eq19}. As shown in Table \ref{table1} and Table \ref{table2}, scaffold diversity and uniqueness increased slightly, whereas FCD showed a more noticeable rise.

As reported in previous studies, most corrected molecules were corrected to the intended molecules but did not necessarily represent the original molecules exactly \cite{intro17}. Therefore, instances in which the corrected molecules differed from their original labels were extracted, and their Bemis–Murcko scaffolds \cite{murcko} were compared. The Bemis–Murcko scaffold represents the core scaffold of a molecule, retaining only its ring system and the minimal backbone connecting these rings. It is widely used in medicinal chemistry for structure-activity relationship (SAR) studies and scaffold hopping. Through this comparison, it was indirectly confirmed that ChemFixer preserves at least the fundamental characteristics of molecules during the correction process. Fig. \ref{fig5} shows the Bemis–Murcko scaffolds of two molecules, where molecule (A) is the original label, and molecule (B) is the one corrected by ChemFixer. Considering that two molecules are deemed pharmacologically similar if their Tanimoto coefficient is $\geq$ 0.85 \cite{sctanimoto}, the Tanimoto coefficient for these two is 0.5102, indicating that they were not corrected into the same molecule. However, the fact that scaffold-level similarity is preserved suggests that ChemFixer performs minimal corrections, which can be interpreted as a key factor contributing to the limited changes observed in FCD and SNN.

\begin{table*}[!t]
\caption{KIBA SCORE RANKINGS OF NEW LIGANDS DISCOVERED THROUGH CHEMFIXER AND THEIR ACTIVATED TARGET PROTEINS}
\label{table3}
\centering
\begin{tabular}{@{}c|cl|c|l@{}}
\toprule
               & \multicolumn{2}{c|}{\begin{tabular}[c]{@{}c@{}}PubChem\\ CID\end{tabular}} & \begin{tabular}[c]{@{}c@{}}Drug SMILES strings\\ (based on PubChem)\end{tabular}         & \multicolumn{1}{c}{\begin{tabular}[c]{@{}c@{}}Activate target\\ (KIBA score ranking \cite{intro18})\end{tabular}}                                                                                                                                                                                                                                                                \\ \midrule
Input drug     & \multicolumn{2}{c|}{49830445}                                              & \begin{tabular}[c]{@{}c@{}}C\underline{C}S(=O)(=O)NC1=CC=C(C=C1)\\ C2=CC3=C(C=C2)NN=C3N\end{tabular} & \begin{tabular}[c]{@{}l@{}}1. DAPK3-Death-associated protein kinase 3 (human)\\ 2. CLK4-Dual specificity protein kinase CLK4 (human)\\ 3. FYN-Tyrosine-protein kinase Fyn (human)\\ 4. PKN2-Serine/threonine-protein kinase N2 (human)\\ …\end{tabular}                                                                                                \\
Corrected drug & \multicolumn{2}{c|}{-}                                                     & \begin{tabular}[c]{@{}c@{}}C\underline{S}S(=O)(=O)NC1=CC=C(C=C1)\\ C2=CC3=C(C=C2)NN=C3N\end{tabular} & \begin{tabular}[c]{@{}l@{}}1. DAPK3-Death-associated protein kinase 3 (human)\\ 2. CLK4-Dual specificity protein kinase CLK4 (human)\\ \textbf{3. DYRK1B-Dual specificity tyrosine-phosphorylation-regulated kinase 1B (human)}\\ \textbf{4. KS6A4-Ribosomal protein S6 kinase alpha-4 (human)}\\ …\end{tabular}                                                         \\ \midrule
Input drug     & \multicolumn{2}{c|}{49830316}                                              & \begin{tabular}[c]{@{}c@{}}C1=CC2=C(NC=C2C3=\\ \underline{C}C(=NC=N3)Cl)N=C1\end{tabular}            & \begin{tabular}[c]{@{}l@{}}1. DYR1A-Dual specificity tyrosine-phosphorylation-regulated kinase 1A (human)\\ 2. CLK4-Dual specificity protein kinase CLK4 (human)\\ 3. CDC7-Cell division cycle 7-related protein kinase (human)\\ 4. CLK2-Dual specificity protein kinase CLK2 (human)\\ …\end{tabular}                                                \\
Corrected drug & \multicolumn{2}{c|}{-}                                                     & \begin{tabular}[c]{@{}c@{}}C1=CC2=C(NC=C2C3=\\ \underline{N}C(=NC=N3)Cl)N=C1\end{tabular}            & \begin{tabular}[c]{@{}l@{}}1. DYR1A-Dual specificity tyrosine-phosphorylation-regulated kinase 1A (human)\\ 2. CLK4-Dual specificity protein kinase CLK4 (human)\\ \textbf{3. DYR1B-Dual specificity tyrosine-phosphorylation-regulated kinase 1B (human)}\\ \textbf{4. KCC2B-Calcium/calmodulin-dependent protein kinase type II subunit beta (human)}\\ …\end{tabular} \\ \bottomrule
\end{tabular}
\end{table*}

\begin{figure*}[!t]
    \centering
    \includegraphics[width=\textwidth]{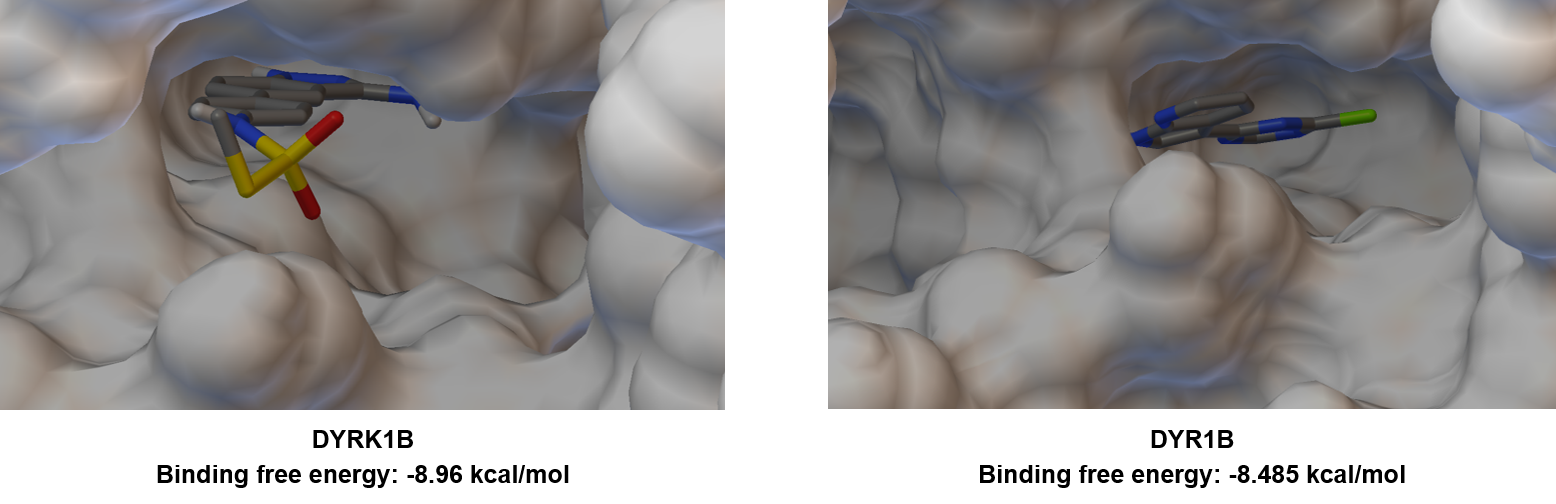} 
    \caption{Estimation of binding free energy and docking image using the AutoDock VINA: ligand–protein pair ranked in the KIBA score in Table \ref{table3}.}
    \label{fig7}
\end{figure*}

\subsection{ChemFixer Results and Correction Effects}
Table \ref{table2} presents the results of sampling 30,000 molecules from each model and correcting the invalid molecules using ChemFixer. ChemFixer improved the validity not only in the base generative model but also across five additional generators that share the MOSES domain but were not used during training. This demonstrates the generalization capability of ChemFixer within the same domain. 

The uniqueness metric increased slightly after correction. Consistent with the observations in Table \ref{table1}, an overall pattern was observed: a modest increase in FCD, a slight decrease in SNN, and a small improvement in molecular diversity (Scaf/TestSF and Unique\text{@}10K). These results suggest that ChemFixer preserves the learned chemical distribution while recovering regions of chemical space previously inaccessible due to decoding errors, indicating that it does not merely regenerate duplicate molecules to boost validity.

Despite the high validity demonstrated by existing molecular generation models, some of the generated molecules still remained invalid. Moreover, the validity metric only checks whether a generated molecule is valid in comparison with its SMILES string, without verifying whether it matches the intended label. This may result in an overestimated value compared with actual expectations. Therefore, it was necessary to confirm whether the generated molecules truly aligned with their correct labels. To investigate this, molecules were generated by conditioning on different values of molecular properties (logP, tPSA, QED). As a result, it was observed that validity varied according to the chosen condition, and many duplicate molecules were produced. These observations indicate that the validity metric does not consistently guarantee high validity under various conditions, and that some molecules remain undecoded. Therefore, we deemed it necessary to experimentally assess whether these molecules could be recovered using ChemFixer.

Accordingly, 5,000 molecules were sampled by specifying the molecular property values as conditions. Fig. \ref{fig6} shows the target molecule generation outcomes for both the baseline model and MolGCT \cite{intro11}: the generated molecules (green circles) and those that were invalid but subsequently corrected by ChemFixer (red/blue triangles). These results yield two important implications. First, ChemFixer performs minimal corrections without over-modifying the molecular structure, thereby preserving the original chemical integrity. Second, through such minimal corrections, ChemFixer successfully recovers molecules that the original model failed to decode during the target-specific generation process, effectively expanding the accessible chemical space. In summary, ChemFixer broadens the range of candidate molecules, thereby supporting a core objective of molecular generative models: enabling the discovery of diverse and novel drug candidates.

\subsection{Application to the Drug–Target Interaction Task}
To evaluate the extensibility of ChemFixer, we applied it to the Co-VAE \cite{intro18} model, which performs a Drug–Target Interaction (DTI) task. Among the 2,111 ligands used to train Co-VAE, 1,998 ligands that satisfied the maximum length requirement defined by ChemFixer were selected for evaluation. Due to the low validity of the molecules generated by Co-VAE, valid/invalid ligand pairs were constructed in more than 65 out of the 100 training epochs, following the same method described in \eqref{eq12} and \eqref{eq13}. These pairs were then used to fine-tune the ChemFixer model, which had been pre-trained as described in \eqref{eq14}–\eqref{eq16}, using the same procedure outlined in \eqref{eq17}. As a result, ChemFixer increased the validity of ligands generated by Co-VAE on the KIBA dataset \cite{KIBA1}, \cite{KIBA2} from approximately 55\% to 87\% by correcting previously invalid ligands, thereby enhancing the potential to discover protein–ligand pairs with possible activity.

After discovering new valid ligands structurally similar to the original ones through ChemFixer, we compared the KIBA scores predicted by Co-VAE for their respective target proteins. Table \ref{table3} shows the KIBA score rankings of target proteins for the original input ligands (Input drug) and the previously invalid ligands corrected by ChemFixer (Corrected drug). Notably, the same proteins occupy the first and second ranks in both ligand sets, while different third- and fourth-ranked target proteins were observed for each. This suggests that the corrected ligands preserved the topological similarity of the original structures while inducing new interactions through subtle modifications in substituents or functional groups, potentially enabling different target binding and pharmacological effects.

To further validate the binding potential at the 3D structural level, we conducted docking simulations using AutoDock VINA \cite{autodock} between two corrected ligands (selected from Table \ref{table3}) and their predicted third-ranked target proteins based on the KIBA scores. As shown in Fig \ref{fig7}, both ligands exhibited stable binding, with binding free energy values below –7.0 kcal/mol, meeting the widely accepted threshold for potential bioactivity \cite{vina1}, \cite{vina2}. These results demonstrate that ChemFixer not only restores the validity of molecular structures but also expands the chemical diversity of active ligands, suggesting its promise as a valuable tool in drug discovery pipelines.

\subsection{Ablation Study}
\subsubsection{Masking Probability and Strategy}

In the pre-training phase of ChemFixer, we conducted a comparative study to determine the optimal masking probability by evaluating five different masking ratios: 5\%, 10\%, 15\%, 20\%, and 30\%. For each configuration, we input 30,000 invalid SMILES sequences into the model and assessed its reconstruction performance. The corresponding results are summarized in Table \ref{table4}. The results showed that a masking probability of 10\% yielded the best performance. This setting most closely aligns with the typical number of errors (1 to 5 tokens per sequence) observed in invalid SMILES from the MOSES dataset. This observation shows a similar pattern to that reported in previous work \cite{intro17}, where varying numbers of artificial errors were inserted into individual sequences to assess model performance. Given the rigid syntax of SMILES—where even a single incorrect token can render the entire sequence invalid—a conservative masking rate of 10\% can be interpreted as a reasonable and effective choice.

We also considered more structured masking strategies, such as masking specific tokens \cite{selective} or continuous spans \cite{span}. However, SMILES errors are not concentrated in particular positions or token types, but are rather distributed throughout the sequence. Additionally, the short semantic units and strong positional dependency in SMILES make it difficult to define meaningful structured masking strategies without introducing ambiguity or subjectivity. Considering these factors, we ultimately adopted a random masking strategy, which offers a more generalizable and unbiased training signal.

\begin{table}[!t]
\caption{PERFORMANCE COMPARISON AT DIFFERENT MASKING RATIOS}
\label{table4}
\centering
\footnotesize
\begin{tabular*}{\columnwidth}{@{\extracolsep{\fill}} cccccc}
\toprule
Evaluation & \multicolumn{5}{c}{Masking Ratio} \\ \cmidrule(lr){2-6}
Metrics (\%) & 5\% & 10\% & 15\% & 20\% & 30\% \\
\midrule
Validity  & 93.89 & 97.18 & 96.31 & 91.17 & 84.23 \\
Precision & 86.27 & 89.73 & 91.37 & 94.33 & 93.79 \\
Recall    & 92.99 & 96.87 & 95.98 & 90.69 & 83.36 \\
F1 Score  & 89.50 & 93.16 & 93.62 & 92.47 & 88.27 \\
\bottomrule
\end{tabular*}
\end{table}

\subsubsection{Effect of Masked Pre-training}
We first evaluated the effect of masked pre-training by applying ChemFixer without masked pre-training to five molecular generative models that share the MOSES domain but were not used during ChemFixer’s training. Even without masked pre-training, ChemFixer achieved an average correction rate improvement of 56\%, suggesting that our proposed data collection and training pipeline is inherently effective. However, this result was approximately 6\% lower than that of the pre-trained version (Table \ref{table2}), indicating that masked pre-training provides meaningful benefits even within the same chemical space.

Next, we applied ChemFixer without masked pre-training to the Co-VAE \cite{intro18} model trained on the KIBA dataset \cite{KIBA1}, \cite{KIBA2}, which differs from MOSES in both chemical distribution and data scale. While the original Co-VAE output achieved a validity of approximately 55\%, ChemFixer without masked pre-training improved this to 66\%. However, this was still about 20\% lower than the 87\% achieved by the pretrained ChemFixer, highlighting that the effect of pre-training becomes even more pronounced in low-resource or distribution-shifted environments.

We attribute this performance gap to the ability of masked pre-training to internalize general SMILES grammar and error patterns—such as broken ring closures and mismatched brackets—from a large-scale corpus, and to the attention mechanism’s ability to effectively capture the conditional probability distribution between error types and correct tokens. With this prior knowledge, the pre-trained model can adapt to a new domain like KIBA dataset even with a small number of invalid–valid molecular pairs. In contrast, a model without masked pre-training must learn the same error patterns from scratch using only limited data, which may pose limitations in terms of learning efficiency and overall performance.

In summary, masked pre-training serves as a critical component of ChemFixer, enabling robust performance gains not only within the same dataset but also under data-scarce or chemically distinct conditions, thereby supporting its potential for broader applicability and generalization.

\section{DISCUSSION AND CONCLUSION}
In this paper, we proposed ChemFixer, a framework designed to correct invalid SMILES generated by deep learning–based molecular generative models. We fine-tuned ChemFixer, which had been pre-trained with a 10\% masking probability, using a dataset of valid/invalid molecular pairs collected from the base Transformer-VAE model. As a result, ChemFixer improved the validity of not only the base generative model but also five additional generators unseen during training, thereby demonstrating generalization performance. Furthermore, fine-tuning the pre-trained ChemFixer on the Co-VAE model \cite{intro18}, which uses data from a chemically distinct domain, greatly improved the validity and led to the discovery of promising protein–ligand pairs.

The contribution of pre-training was confirmed through ablation studies. When the masking-based pre-training stage was omitted, the correction rate dropped by approximately 6\% in the MOSES domain and by about 20\% in the low-resource DTI task. This suggests that pre-training on large-scale corpora provides a strong inductive bias by learning SMILES syntax and error patterns in advance, which is particularly beneficial in low-data or distribution-shifted scenarios.

Validity is only meaningful when the corrected molecules remain within the drug-like chemical space learned by the generative models. If ChemFixer were to over-correct invalid SMILES into unrealistic structures, the validity score could be artificially inflated while the utility of the generated molecules in downstream tasks such as docking could deteriorate. To assess this risk, we conducted quantitative evaluations using distributional alignment metrics. We observed a consistent pattern of modest increases in FCD, slight decreases in SNN, and small improvements in molecular diversity, indicating that ChemFixer preserves the learned chemical distribution while recovering regions that were previously inaccessible due to decoding errors. These findings align with the validity–diversity–distributional stability trade-offs highlighted in recent studies \cite{dist1}, \cite{dist2}.

Nonetheless, this study is limited by the use of the MOSES token vocabulary, which constrains the ability to process long and complex sequences involving stereochemistry and macrocyclic structures. This limitation suggests that ChemFixer may still present limitations for application in real-world drug design settings.
In future work, we plan to incorporate a 3D conformer-based validation pipeline to better handle stereochemically complex molecules and to extend the applicability of ChemFixer to various stages of drug development.

\section*{References}

\bibliographystyle{ieeetr}
\bibliography{ref}

\end{document}